\documentclass[utf8]{article} 
\usepackage{arxiv}
\usepackage{url,hyperref,lineno,microtype,subcaption}
\usepackage{times}
\usepackage{epsfig}
\usepackage{graphicx}
\usepackage{amsmath}
\usepackage{amssymb}
\usepackage{multirow}
\usepackage{booktabs}
\usepackage{enumitem}
\usepackage{xcolor}
\usepackage{comment}
\usepackage{pgfplots}
\usepackage{standalone}

\usepackage{amsmath,amsfonts,bm, amsthm,algorithm, algpseudocode}

\def\Figref#1{Fig.~\ref{#1}}

\def\eqref#1{equation~\ref{#1}}

\def\Algref#1{Alg.~\ref{#1}}

\def\Tableref#1{Table~\ref{#1}}

\def\1{\bm{1}}

\def\rvx{{\mathbf{x}}}
\def\rvy{{\mathbf{y}}}

\def\rmW{{\mathbf{W}}}

\def\vx{{\bm{x}}}

\DeclareMathAlphabet{\mathsfit}{\encodingdefault}{\sfdefault}{m}{sl}
\SetMathAlphabet{\mathsfit}{bold}{\encodingdefault}{\sfdefault}{bx}{n}

\def\sR{{\mathbb{R}}}

\theoremstyle{plain}

\theoremstyle{remark}

\theoremstyle{definition}

\theoremstyle{plain}

\theoremstyle{plain}

\theoremstyle{definition}

\providecommand{\corollaryname}{Corollary}
\providecommand{\lemmaname}{Lemma}
\providecommand{\problemname}{Problem}
\providecommand{\remarkname}{Remark}
\providecommand{\theoremname}{Theorem}

\pgfplotsset{compat=1.16}

\title{Adversarial Token Attacks on Vision Transformers}
\author{Ameya Joshi\qquad Gauri Jagatap \qquad Chinmay Hegde\\
New York University\\
\tt{\small{\{ameya.joshi, gbj221, chinmay.h\}@nyu.edu}}}
\begin{document}




\maketitle

\begin{abstract}

Vision transformers rely on a patch token based self attention mechanism, in contrast to convolutional networks. We investigate fundamental differences between these two families of models, by designing a block sparsity based adversarial token attack. We probe and analyze transformer as well as convolutional models with token attacks of varying patch sizes. We infer that transformer models are more sensitive to token attacks than convolutional models, with ResNets outperforming Transformer models by up to $\sim30\%$ in robust accuracy for single token attacks.

\end{abstract}


\section{Introduction}
\label{sec:intro}

\subsection{Motivation} Convolutional networks (CNNs) have shown near human performance in image classification~\cite{Kolesnikov2020BigT} over non-structured dense networks. 
However, CNNs are vulnerable to specifically designed adversarial attacks~\cite{adversarialexamples2015}.  Several papers in adversarial machine learning literature reveal the brittleness of convolutional networks to adversarial examples. For example, gradient based methods \cite{Goodfellow2018existence,Kurakin2017} design a perturbation by taking steps proportional to the gradient of the loss of the input image $x$ in a given $\ell_p$ neighborhood. 
This has led to refined robust training approaches, or defenses, which train the network to see adversarial examples during the training stage and produce the unaltered label corresponding to it \cite{madry2018towards,trades}.

Vision transformers (ViT) were recently introduced \cite{Dosovitskiy2021AnII}, as a new network architecture inspired by transformers \cite{Vaswani2017AttentionIA} which have been successfully used for modeling language data. ViTs rely on self attention \cite{Vaswani2017AttentionIA}, a mechanism that allows the network to find correlations between spatially separated parts of the input data. In the context of vision, these are small non-overlapping \textit{patches} which serve as \textit{tokens} to the transformer. ViTs and more recently distillation based Data Efficient Image Transformers (DeIT) \cite{Touvron2021TrainingDI} have shown to have competitive performance on classification tasks and rely on pre-training on very large datasets. It is of imminent interest to therefore study the robustness of self-attention based networks.

There has been some preliminary work on adversarial robustness of vision transformers. \cite{Bhojanapalli2021UnderstandingRO} show that under certain regimes, vision transformers are at least as robust to $\ell_2$ and $\ell_\infty$ PGD attacks as ResNets. While $\ell_2$ and $\ell_\infty$ threat models are useful in understanding fundamental properties of deep networks, they are not realizable in the real world and do not capture actual threats. Transformer based networks also introduce the need for tokenizing the image, leading to an encoded bias in the input. It is therefore important to understand the sensitivity of the architecture to token level changes rather than to the full image.

Specifically, we attempt to answer:
\emph{Are transformers robust to perturbations to a subset of the input tokens?} We present a systemic approach to answer this query by constructing token level attacks by leveraging block sparsity constraints. 

\subsection{Our contributions} 

In this paper, we propose a patch based block sparse attack where the attack budget is defined by the number of tokens the attacker is allowed to perturb. We identify top salient pixels using the magnitude of their loss gradients and perturb them to create attacks. We extend a similar idea to block sparsity by constraining salient pixels to lie in non-overlapping patches. We probe three families of neural architectures using our token attack; self-attention (ViT~\cite{Dosovitskiy2021AnII}, DeIT~\cite{Touvron2021TrainingDI}), convolutional (Resnets~\cite{He2016DeepRL} and WideResNet~\cite{Zagoruyko2016WideRN}) and MLP based (MLP Mixer~\cite{Tolstikhin2021MLPMixerAA}).

We make the following contributions and observations:
\begin{enumerate}[nolistsep, left=0pt]
\item We propose a new attack which imposes block sparsity constraints, allowing for \textit{token attacks} for Transformers. 

\item We show classification performance of all architectures on token attacks of varying patch sizes and number of patches.

\item We demonstrate that for token attacks matching the architecture token size, vision transformers are less resilient to token attacks as compared to MLP Mixers and ResNets.

\item For token attacks smaller than architecture token size, vision transformers are comparably robust to ResNets.

\item We also specifically note the shortcomings of previous studies on robustness of transformers~\cite{Bhojanapalli2021UnderstandingRO}, where ViTs are shown to be more robust than ResNets.  

\item With our token attacks we can break Vision transformers using only $1\%$ of pixels as opposed to $\ell_2$ or $\ell_\infty$ attacks which rely on perturbing all image pixels.

\end{enumerate}
We therefore motivate designing attacks adaptively modeled after neural architectures.

\subsection{Related work}

\noindent\textbf{\textit{Threat models:}}  Deep networks are vulnerable to imperceptible changes to input images as defined by the $\ell_\infty$ distance \cite{Szegedy2014intriguing}. There exist several test-time attack algorithms with various threat models: $\ell_p$ constrained~\cite{adversarialexamples2015, Kurakin2017, Carlini2017cwl2}, black-box~\cite{Ilyas2018blackbox, Ilyas2018alimitedqueries}, geometric attacks~\cite{engstrom2019a, Xiao2018SpatiallyTA}, semantic and meaningful attacks~\cite{joshi2019semantic, zhang2019camou, song2018constructing} and data poisoning based~\cite{Shafahi2018poisonfrogs}.

\noindent\textbf{\textit{Defenses:}} Due to the vast variety of attacks, adversarial defense is a non-trivial problem. Empirical defenses as proposed by \cite{madry2018towards}, \cite{trades}, and \cite{jagatap2020adversarially} rely on adversarial data augmentation and modified loss functions to improve robustness. Several works~\cite{samangouei2018defensegan, yin2020defense} propose preprocessing operations as defenses. However, such defenses often fail to counter adaptive attacks~\cite{Athalye2018obfuscated}. \cite{wong2018provable}, \cite{cohen2019certified} and \cite{salman2019provably} provide methods that guarantee robustness in terms of a volume around an input. Such methods often fail or provide trivial certificates for larger networks, and large high resolution images. Apart from algorithmic approaches, newer papers discuss optimal hyper-parameter tuning as well as combination of regularizers from aformentioned techniques, choice of activation functions, choice of architecture and data augmentation to extract best possible robust accuracies using pre-existing algorithms \cite{Gowal2020UncoveringTL, Pang2021BagOT}.

\noindent\textbf{\textit{Patch attacks:}} Patch attacks~\cite{Brown2017advpatch} are practically realizable threat model. \cite{zolfi2021translucent, thys2019fooling, wu2020making} have successfully attacked detectors and classifiers with physically printed patches. In addition, \cite{croce2019sparse, croce2019sparse} also show that spatially limited sparse perturbations suffice to consistently reduce the accuracy of classification model. This motivates our analysis of the robustness of recently invented architectures towards sparse and patch attacks.



\noindent\textbf{\textit{Vision transformers}}
While convolutional networks have successfully achieved near human accuracy on massive datasets~\cite{Kolesnikov2020BigT, Xie2020SelfTrainingWN}, there has been a surge of interest in leveraging self-attention as an alternative approach. Transformers~\cite{Vaswani2017AttentionIA} have been shown to be extremely successful at language tasks~\cite{Devlin2019BERTPO, Sanh2019DistilBERTAD, Brown2020LanguageMA}. \cite{parmar2018image} extend this for image data, where in they use pixels as tokens. While they some success in generative tasks, the models had a large number of parameters and did not scale well. \cite{Dosovitskiy2021AnII} improve upon this by instead using non-overlapping patches as tokens and show state of the art classification performance on the ImageNet dataset. \cite{Touvron2021TrainingDI} further leverage knowledge distillation to improve efficiency and performance. Further improvements have been suggested by \cite{Dai2021CoAtNetMC}, \cite{Wu2021CvTIC} and \cite{Touvron2021GoingDW} to improve performance using architectural modifications, deeper networks and better training methods. In parallel, \cite{Tolstikhin2021MLPMixerAA} instead propose a pure MLP based architecture that achieves nearly equivalent results with faster training time. However, studies on generalization and robust performance of such networks is still limited. We discuss a few recent works below.

\noindent\textbf{\textit{Attacks on vision transformers:}}
\cite{Bhojanapalli2021UnderstandingRO,hendrycks2020pretrained}  analyse the performance of vision transformers in comparison to massive ResNets under various threat models and concur that vision transformers (ViT) are at least as robust as Resnets when pretrained with massive training datasets. \cite{mahmood2021robustness} show that adversarial examples do not transfer well between CNNs and transformers, and build an ensemble based approach towards adversarial defense. \cite{paul2021vision} claims that Transformers are robust to a large variety of corruptions due to attention mechanism.

\section{Token Attacks on Vision transformers}
\label{sec:blocksparse}


\noindent\textbf{Threat Model:} We define the specific threat model that we consider in our analysis. Let $\rvx \in \sR^d$ be a $d$-dimensional image, and $f:\sR^d \to [m]$ be a classifier that takes $\rvx$ as input and outputs one of $m$ class labels. For our attacks, we focus on sparsity as the constraining factor. Specifically, we restrict the number of pixels or blocks of pixels that an attacker is allowed to change. We consider $\rvx$ as a concatenation of $B$ blocks $[\vx_1, \dots \vx_b, \dots, \vx_B]$, where each block is of size $p$. In order to construct an attack, the attacker is allowed to perturb up to $K\leq B$ such blocks for a $K$-token attack. We also assume a white-box threat model, that is, the attacker has access to all knowledge about the model including gradients and preprocessing. We consider two varying attack budgets. In both cases we consider a block sparse token budget, where we restrict the attacker to modifying $K$ patches or ``tokens" (1) with an unconstrained perturbation allowed per patch (2) a ``mixed norm'' block sparse budget, where the pixelwise perturbation for each token is restricted to an $\ell_\infty$ ball with radius $\epsilon$ defined as $K, \epsilon$-attack. 


\noindent\textbf{Sparse attack:}
To begin, consider the simpler case of a sparse ($\ell_0$) attack. This is a special case of the block sparse attack with block size is \emph{one}. Numerous such attacks have been proposed in the past (refer to appendix). The general idea behind most such attacks is to analyse which pixels in the input image tend to affect the output the most 
    $S(x_{i}) := \left |\frac{\partial L(f(\rvx, \rvy))}{\partial x_{i}}\right|$,
where $L(\cdot)$ is the adversarial loss, and $c$ is the true class predicted by the network. 
The next step is to perturb the top $s$ most salient pixels for a $s$-sparse attack by using gradient descent to create the least amount of change in the $s$ pixels to adversarially flip the label.

\noindent\textbf{Patchwise token attacks:}Instead of inspecting saliency of single pixel we check the norm of gradients of pixels belonging to non-overlapping patches using patch saliency $S(\mathbf{x}_b) := \sqrt{\sum_{x_i\in \vx_b} \left |\frac{\partial L(f(\rvx, \rvy))}{\partial x_{i}}\right|^2 }$, for all $b\in \{1,\dots B\}$. We pick top $K$ blocks according to patch saliency. The effective sparsity is thus $s=K\cdot p$.
These sequence of operations are summarized in \Algref{alg:attack}. 

\begin{algorithm}[tp]
\small
\caption{Adversarial Token Attack}
\label{alg:attack}
\begin{algorithmic}[1]
\Require  $\rvx_0$:Input image, $f(.)$: Classifier, $\rvy:$ Original label, $K$: Number of patches to be perturbed, $p$: Patch size.
 $i \gets 0$
\State $[b_1\dots b_K]$= Top-K of $S(\mathbf{x}_b) = \sqrt{\sum_{x_i\in \vx_b} \left |\frac{\partial L(f(\rvx, \rvy))}{\partial x_{i}}\right|^2 }$,~$\forall b$.
\While $f(\rvx) \neq y \text{ OR MaxIter}$
    \State $\rvx_{b_k} = \rvx_{b_k} + \nabla_{\rvx_{b_k}} L;~~\forall~~b_k~\in~\{b_1,\dots, b_K\}$
    \State $ \rvx_{b_k} = Project_{\epsilon_\infty} (\rvx_{b_k})$ (optional)
\EndWhile
\end{algorithmic}
\end{algorithm}

We use non-overlapping patches to understand the effect of manipulating salient tokens instead of arbitrarily choosing patches. In order to further test the robustness of transformers, we also propose to look at the minimum number of patches that would required to be perturbed by an attacker. For this setup, we modify \Algref{alg:attack} by linearly searching over the range of $1$ to $K$ patches.

\begin{figure*}
\centering
\begin{tabular}{c}
 \hspace{15pt} Original \hspace{38pt} Adversarial (patch) \hspace{8pt} Pertubation (patch) \hspace{10pt} Adversarial (sparse) \hspace{8pt} Pertubation (sparse)\\
 \includegraphics[height=0.15\linewidth]{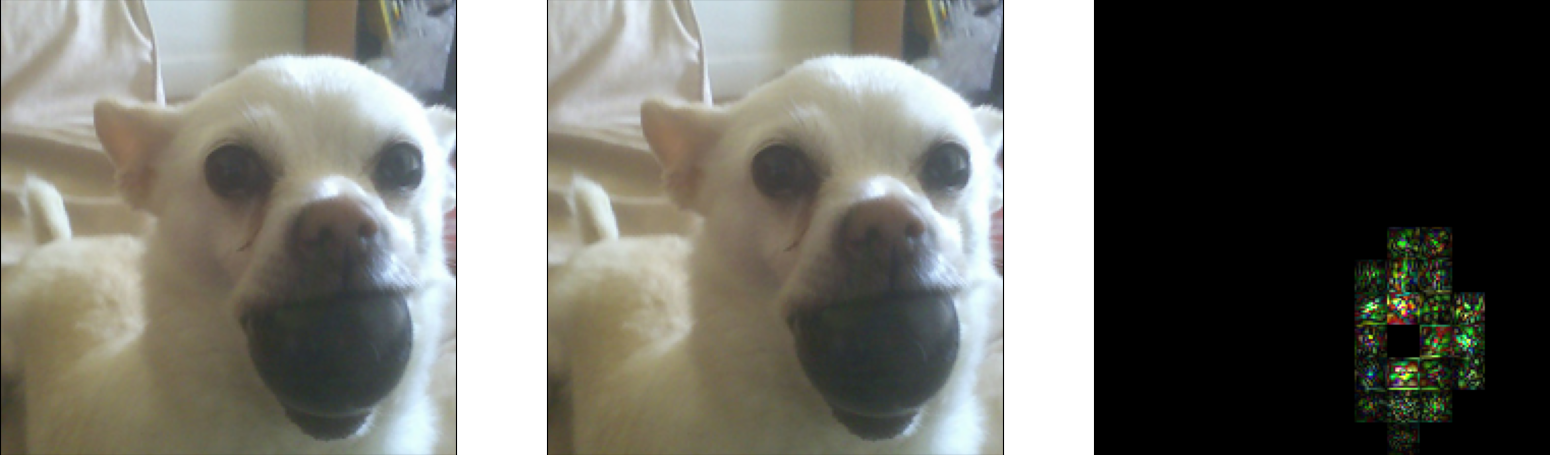} \includegraphics[height=0.15\linewidth]{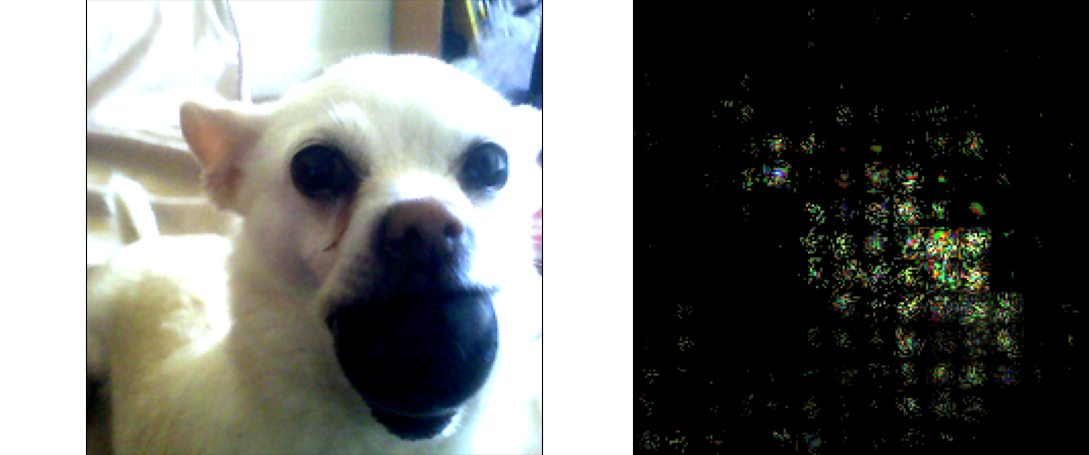}\\
\end{tabular}

\caption{\sl\textbf{Patch and sparse attacks on transformers}: The attack images are generated with a fixed budget of $20$ patches of size $16\times 16$, or $5120$ pixels for sparse attack on vision transformer (ViT). Note that the perturbations are imperceptible. The third and fifth columns shows the perturbations brightened $10$ times.}
\label{fig:vis_blocksparse}
\end{figure*}

\noindent\textbf{{Mixed-norm attacks:}}
Most approaches~\cite{croce2019sparse, croce2020sparse} additionally rely on a mixed $\ell_2$-norm based sparse attack in order to generate imperceptible perturbations. Motivated by this setting, we propose a mixed-norm version of our modified attack as well. In order to ensure that our block sparse attacks are imperceptible, we enforce an additional $\ell_\infty$ projection step post the gradient ascent step. This is enforced via Step 4 in Alg. \ref{alg:attack}.







\section{Experiments and Results}
\label{sec:exps}

\noindent\textbf{Setup:}  To ensure a fair comparison, we choose the best models for the Imagenet dataset~\cite{ILSVRC15} reported in \cite{Dosovitskiy2021AnII}, \cite{Touvron2021TrainingDI} and \cite{Zagoruyko2016WideRN}. The models achieve near state-of-the-art results in terms of classification accuracy. They also are all trained using the best possible hyperparameters for each case. We use these weights and the shared models from the \texttt{Pytorch Image models}~\cite{rw2019timm} repository. 
We restrict our analysis to a fixed subset of $300$ randomly chosen images from the Imagenet validation dataset. 

\noindent\textbf{Models:} In order to compare the robustness of transformer models to standard CNNs, we consider three different families of architectures:(1) Vision Transformer (ViT)~\cite{Dosovitskiy2021AnII}, Distilled Vision Transformers (DeIT)~\cite{Touvron2021TrainingDI}, (2) Resnets~\cite{He2016DeepRL, Zagoruyko2016WideRN} and (3) MLP Mixer \cite{Tolstikhin2021MLPMixerAA}. 
For transformers, \cite{Dosovitskiy2021AnII} show that best performing Imagenet models have a fixed input token size of $16\times16$.
In order to ensure that the attacks are equivalent, we ensure that any norm or patch budgets are appropriately scaled as per the pre-processing used \footnote{In case of varying image sizes due to pre-processing, we calculate the scaling factor in terms of the number of pixels and appropriately increase or decrease the maximum number of patches.}. We also scale the $\epsilon$-norm budget for mixed norm attacks to eight gray levels of the input image post normalization. Additionally, we do a hyper parameter search to find the best attacks for each model analysed. Specific details can be found in the Appendix\footnote{\url{https://github.com/NYU-DICE-Lab/TokenAttacks_Supplementary.git}}.

\begin{figure*}[htp]
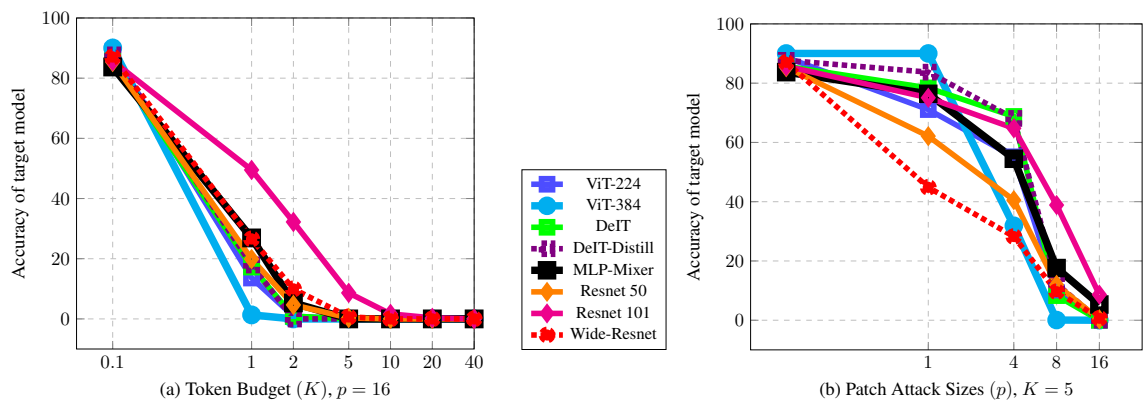

    \centering
    \begin{tabular}{c c }
         \includegraphics[width=0.39\linewidth]{plots/varypatches_16.tex} &
         \includegraphics[width=0.5\linewidth]{plots/varyingtokens.tex} 
    \end{tabular}
    \caption{(a) \small \sl \textbf{Robustness to Token Attacks with varying budgets ($p=16$).} Vision transformers are less robust than MLP Mixer and ResNets against patch attacks with patch size matching token size of transformer architecture, (b) \textbf{Token attacks with varying patch sizes.$K=5$} When the attack patch size is smaller than token size of architecture, vision transformers are comparably robust against patch attacks, to MLP and ResNets. Detailed results can be found in the Appendix }
    \label{fig:results}
\end{figure*}

\noindent\textbf{Patch attacks:} We allow the attacker a fixed budget of tokens as per Algorithm \ref{alg:attack}. 
We use the robust accuracy as the metric of robustness, where a higher value is better.
We start with an attack budget of $1$ token for an image size of $224\times224$ for the attacker where each token is a patch of the size $16\times 16$. In order to compensate for the differences in the size of the input, we scale the attack budget for ViT-384 by allowing for more patches ($3$ to be precise) to be perturbed.  However, we do not enforce any imperceptibility constraints. We run the attack on the fixed subset of ImageNet for the network architectures defined above.  \Figref{fig:results}(a) shows the result of our analysis. Notice that Transformer architectures are more vulnerable to token attacks as compared to ResNets and MLP-Mixer. Further, ViT-384 proves to be the most vulnerable, and ResNet-101 is the most robust model. DeiT which uses a teacher-student network is more robust than ViTs. We therefore conclude that distillation improves robustness to single token attacks. 

\noindent\textbf{\textit{Varying the Token budget:}} For this experiment, we start with a block-budget of $1$ patch, and iterate upto $40$ patches to find the minimum number of tokens required to break an image. We then measure the robust accuracy for each constraint and for each model. For this case, we only study attacks for a fixed patch (token) size of $16\times16$ and represent our findings in \Figref{fig:results}(a). We clearly observe a difference in the behavior of ViT versus ResNets here. In general, for a given token budget, ResNets outperform all other token based models. In addition, the robust accuracies for Transformers fall to zero for as few as \emph{two} patches. The advantage offered by distillation for single token attacks is also lost once the token budget increases. 

\noindent\textbf{\textit{Varying patch sizes:}}In order to further analyse if these results hold across stronger and weaker block sparse constraints, we further run attacks for varying patch sizes. Smaller patch sizes are equivalent to partial token manipulation. We fix the token budget to be $5$ or $15$ tokens as dictated by the input size. Here, this corresponds to allowing the attacker to perturb $5$ $p\times p$ patches. As one would expect, a smaller partial token attack is weaker than a full token attack. Surprisingly, the Transformer networks are comparable or better than ResNets for attacks smaller than a single token. This leads us to conclude that Transformers can compensate for adversarial perturbations within a tokens. However, as the patch size approaches the token size, Resnets achieve better robustness. We also see that MLP-Mixers, while also using the token based input scheme, perform better than Transformers as the patch attack size increases.

However, this approach allows for unrestricted changes to the tokens. Another approach would be to study the effect of ``mixed norm'' attacks which further constrain the patches to be \emph{imperceptibly} perturbed.
\noindent\textit{\textbf{Mixed Norm Attacks:}} For the mixed norm attacks, we analyse the robustness of all networks for a fixed $\epsilon$ $\ell_\infty$ budget of one gray level. We vary the token budgets from $1$ to $5$. Here, almost all the networks show similar robustness for a small token budget ($K$=1,2); refer \Tableref{t:mixednorm}. However, as the token budget increases, Transformer and MLP Mixer networks are far more vulnerable. Note that this behavior contradicts~\cite{Bhojanapalli2021UnderstandingRO}, where ViTs outperform ResNets. Since our threat model leverages the token based architecture of the Transformers, our attacks are far more successful at breaking ViTs over Resnets.
\begin{small}
\begin{table}[h]
\centering
\begin{minipage}{0.48\linewidth}
\caption{\small \textbf{\sl  Robust Accuracy for Mixed Norm Attacks:} The models are attacked with a $K,(1/255)$ Patch Attack. Note that for smaller token budgets, the models perform nearly the same. However, as the token budget increases, Resnets are more robust than Transformers.}
\label{t:mixednorm}
\begin{tabular}{c c c c c} 
  \toprule
  \textbf{Model} & \textbf{Clean} & \multicolumn{3}{c}{\textbf{Token Budget}}  \\
  \midrule 
  {} & & 1 & 2 & 5 \\
  \midrule
   ViT-224 & 88.70 & 68.77 & 50.83 & 15.28 \\
   ViT-384 & \textbf{90.03} & \textit{53.48} & \textit{28.57} & \textit{4.98} \\ 
   \midrule
   DeIT & 85.71 & \textbf{72.42}  & 46.84 & 6.31 \\
   DeIT-Distilled & 87.70 &  68.77 & 54.15 & 16.61 \\
   \midrule
   Resnet-101 & 85.71 & 69.10 & 55.14 & \textbf{32.89} \\
   Resnet-50 & 85.38 &  67.44 & \textbf{55.81} & 31.22 \\
   Wide Resnet & 87.04 &  54.81 & 32.89 & 11.62 \\
   \midrule
   MLP-Mixer & \textit{83.78} &  63.78 & 37.87 & 5.98 \\
  \bottomrule    
\end{tabular}
\end{minipage}
\begin{minipage}{0.48\linewidth}
 \centering
    \small
    \caption{\sl \textbf{Robust accuracies, $s=256$ sparse and $K=1$, $16\times16$ patch attack .} }
    \begin{tabular}{c c c c}
        \toprule
        \textbf{Model} & & \multicolumn{2}{c}{\textbf{Norm constraint}} \\
        \midrule 
        {} & Clean & Sparse  & Patch\\
        \midrule 
        ViT 224 & 88.70 & 5.98 & 13.62 \\
        ViT 384 & \textbf{90.03} & 3.32 & \textit{1.33}\\
        \midrule
        DeIT & 85.71 & 4.65  & 17.27  \\
        DeIT (Distilled) & 87.70& 14.95  &  17.94\\
        \midrule
        MLP Mixer & \textit{83.72} & 5.98 & 26.91 \\
        \midrule
        ResNet 50 & 85.38 & 13.95 & 19.90  \\
        ResNet 101 & 85.71 & \textbf{23.59} & \textbf{49.50} \\
        Wide Resnet & 87.04 & \textit{1.33} & 26.57  \\
        \bottomrule \\
    \end{tabular}   
    \label{tab:sparse}
\end{minipage}
\end{table}
\end{small}

\noindent\textbf{Sparse Attacks:} The sparse variant of our algorithm restricts the patch size to $1\times 1$. We allow for a sparsity budget of $0.5\%$ of original number of pixels. In case of the standard $224\times 224$ ImageNet image, the attacker is allowed to perturb $256$ pixels. We compare the attack success rate of both sparse attack and patch-based token attack at same sparsity budget; to compare we chose $1, 16\times 16$ patch attack (refer \Tableref{tab:sparse}). 
We see that as is the case with token attacks, even for sparse attacks, vision transformers are less robust as compared to ResNets. With the same sparsity budget, sparse attacks are stronger than token attacks; however we stress that sparse threat model is less practical to implement as the sparse coefficients may be scattered anywhere in the image. 
\section{Discussion and Conclusion}
Analysing the above results, we infer certain interesting properties of transformers. 
\begin{enumerate}[nolistsep, parsep=0pt]
 \item We find that Transformers are generally susceptible to token attacks, even for very low token budgets.
 \item However, Transformers appear to compensate for perturbations to patch attacks smaller than the token size. 
 \item  Further, ResNets and MLP-Mixer outperform Transformers for token attacks consistently.
\end{enumerate}

An interesting direction of follow-up work is to develop strong certifiable defenses for token attacks. Further directions of research also include analysis of the effect of distillation and semi-supervised pre-training.
 
\section*{Acknowledgements}
The authors were supported in part by the National Science Foundation under grants CCF-2005804 and CCF-1815101, USDA/NIFA under grant USDA-NIFA:2021-67021-35329, and ARPA-E under grant DE:AR0001215.





\bibliographystyle{IEEEbib} 
\begin{small}
\bibliography{egbib}
\end{small}
\clearpage
\appendix

\section{Background}
\subsection{Transformers}
The Transformer block was introduced by \cite{Vaswani2017AttentionIA}, for text input. The basic idea of the Transformer model is to leverage an efficient form of ``self-attention''. A standard attention block is formally defined as,
\begin{equation}
    \rvx_{out} = \text{Softmax}\left(\frac{\rvx\rmW_Q \rmW_k \rvx^T}{\sqrt{d}}\right) \rvx \rmW_V,
    \label{eq:selfatt}
\end{equation}
where $\rvx \in \sR^{d\times n}$ is an input string, $\rvx_{out} \in \sR^{d\times n}$ is the output of the self-attention block, $\rmW_Q,~\rmW_K~\text{and}\rmW_V$ are the learnable \emph{query}, \emph{key} and the \emph{value} matrices. Note that $\rvx$ is actually a concatenation of $n$ ``tokens'' of size $d$, which each represent some part of the input. 
\emph{Multi-headed self attention} stacks multiple such blocks in a single layer. The Transformer model has multiple such layers followed by a final output attention layer with a \emph{classification token}. This architecture makes perfect sense for text where-in tokens are word or sentence embeddings, and each token therefore holds some semantic meaning. These models are trained in an auto-regressive fashion with additional losses for downstream tasks. 

However, extending the same architecture for images is non-trivial; primarily as the atomic components of an image are pixels which hold little to no meaning by themselves. \cite{parmar2018image} propose a solution where they use pixels as tokens and train generative models to solve problems such as image generation and super-resolution. However, the large dimensionality of images forces the Attention blocks to be massively parameterized, leading to issues of scale. In order to remedy this, \cite{Dosovitskiy2021AnII} suggest using local image patches as tokens. This instantly reduces the number of tokens while also leveraging the local consistency property of images. They find that in most cases, it is enough to use non-overlapping patches of $16\times16$ as tokens to ensure near state of the art accuracies. One disadvantage of such massive models however is the requirement of very large training datasets. \cite{Touvron2021TrainingDI} propose a data-efficient distillation based method to train Transformers. Their architecture (DeIT) leverages a custom transformer based distillation token as well as standard student-teacher training approaches to improve both the sample complexity and the performance over Vision Transformers.  

A standard Resnet model, on the other hand, uses residual blocks:
\begin{equation}
    \rvx_{out} = \text{ReLU}\left(\rvx + \text{ReLU}(\rmW\rvx)\right).
    \label{eq:res_block}
\end{equation}
A Resnet stacks several such residual blocks in succession followed by a classifier. The residual connection allows for easy gradient flow and improves training. There have been several works that prove the generalization and efficacy of Resnets, both empirically~\cite{He2016DeepRL} and theoretically~\cite{Yun2019AreDR}.

\subsection{How resnets differ from transformers}

In comparison with Resnets, which were the best performing image classifiers previously, we see that there are two major structural differences. The first is that most Resnets downsample activations as we go deeper. This is supposed to help reduce redundancies and propagate discriminative features. However, Vision Transformers with self-attention blocks appear to preserve activation sizes throughout their depth. The second major difference is the structure of the Resnet block in comparison with the Attention block. As is evident, any interaction between non-local pixel groups in Resnets is happens in deeper layers. The initial layers tend to just focus on neighbourhood pixel interactions. However, the Attention mechanism forces each layer of the transformer to consider both local and non-local interactions. There exist additional differences in terms of the non-linearities involved and the number of parameters in each model.

The specific difference in the treatment of local and non-local pixel groups informs our choice of attack. While several papers have previously studied the robustness of vision transformers in the standard adversarial setting, we specifically consider the case where the attacker is only allowed to modify an image locally; for example a set number of tokens. 

\subsection{Saliency attacks}
Such `salient' pixels are often identified using the magnitudes of gradients. This idea, while not particularly new~\cite{simonyan2013deep}, lends itself naturally to constructing adversarial attacks. Specifically, the idea is to only perturb a subset of the salient pixels thus implicitly satisfying the sparsity constraint. JSMA~\cite{Papernot2016TheLO} and Maximal-JSMA~\cite{Wiyatno2018MaximalJS} leverage this observation to construct $k$-sparse attacks by maximally perturbing $k$ salient pixels. In maximal-JSMA, the authors calculate saliency of each pixel usign the following equation;
\begin{equation}
    S^{+}(x_{i, c}) = \begin{cases} 
        0~\text{if}~\frac{\partial f(\rvx)_c}{\partial x_i} < 0~\text{or}~\sum_{c'\neq c}\frac{\partial f(\rvx )_c'}{\partial x_i} \\
        -\frac{\partial f(\rvx)_c}{\partial x_i}\cdot \sum_{c'\neq c}\frac{\partial f(\rvx )_c'}{\partial x_i} ~\text{otherwise},\\
        \end{cases}
    \label{eq:jsma_sparse}
\end{equation}
where $x_i$ is the pixel in question, $c$ is the true class, and $f_i$ is a logit value specific to class $i$. 

In this paper, we propose a patch based block sparse attack where the attack budget is defined by the number of patches (blocks) the attacker is allowed to perturb. Our approach  builds on JSMA~\cite{Papernot2016TheLO} Maximal-JSMA~\cite{Wiyatno2018MaximalJS}, wherein the attacker identifies top salient pixels using gradients and perturb them to create attacks. We extend a similar idea to block sparsity. The main differences between JSMA and our approach lie in two places: (1) We use a simplified construction for the saliency map that relies on the magnitude of the gradients with respect to each pixel, (2) instead of considering salient pixels, we instead identify the most informative pixel blocks and further rely on gradient updates to generate an attack. 

\section{Experiments}
For all experiments, we use SGD for optimization with a step size of $0.1$ for a maximum of $100$ steps for both variants for all models except Resnet-101. For Resnet-101, we use a step size of $0.2$ for a maximum of $100$ steps.


\subsection{Mixed norm attacks}


For mixed norm block sparse attacks, we impose an additional $\ell_\infty$ bound $(\epsilon)$ on each pixel to enforce imperceptibility. We run our experiments with a constraint of one gray level similar to \cite{Bhojanapalli2021UnderstandingRO}. Since each of these models scales the input images to varying input ranges, we further scale each $\epsilon$ appropriately. We then use a projection step in \Algref{alg:attack} using clipping to enforce the constraint. 
We show some example attack images in \Figref{fig:attack_mixed}. Thus, we conclude that the attack algorithm implicitly creates imperceptible perturbations.

\section{Detailed Results}

\label{appdx:detailedres}

\begin{figure}[h]
\centering
\begin{tabular}{c c}
Model &  Original \hspace{40pt} Adversarial \hspace{50pt} Pertubation \\
 \raisebox{3em}{ViT384} & \includegraphics[width=0.55\linewidth]{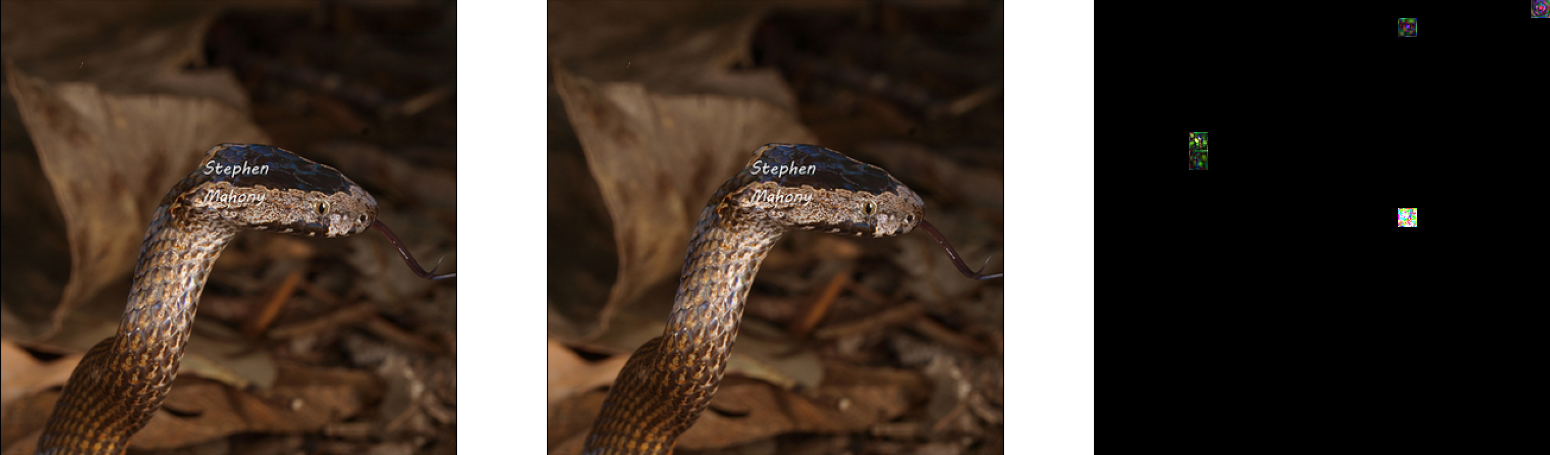} \\ 
 \raisebox{3em}{WideResnet} & \includegraphics[width=0.55\linewidth]{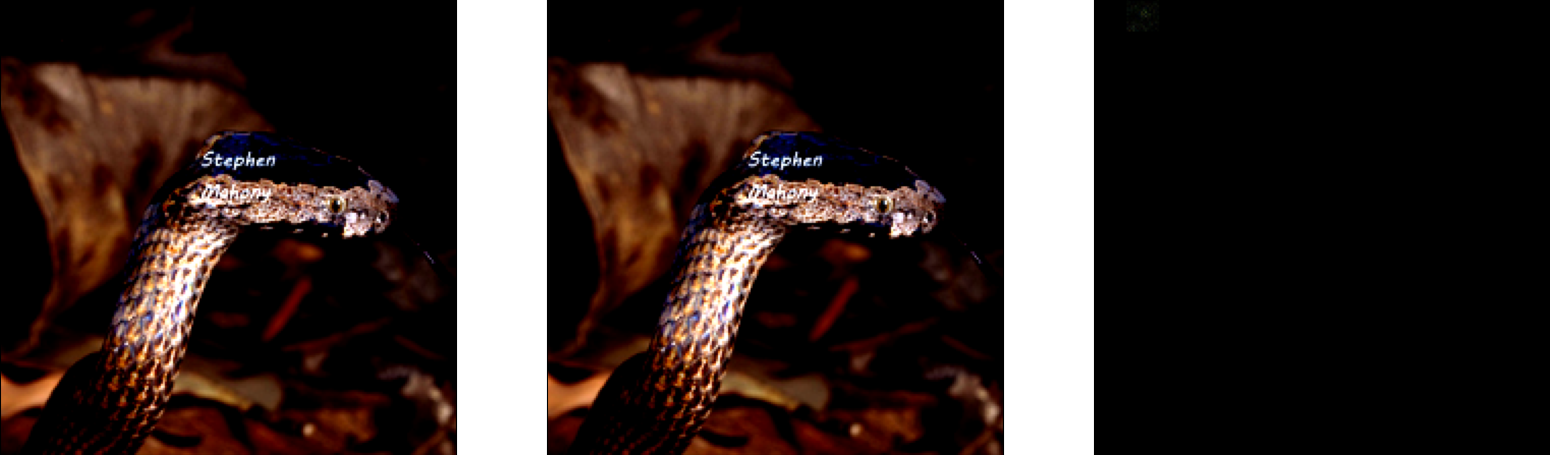} \\
  \raisebox{3em}{DeIT224} & \includegraphics[width=0.55\linewidth]{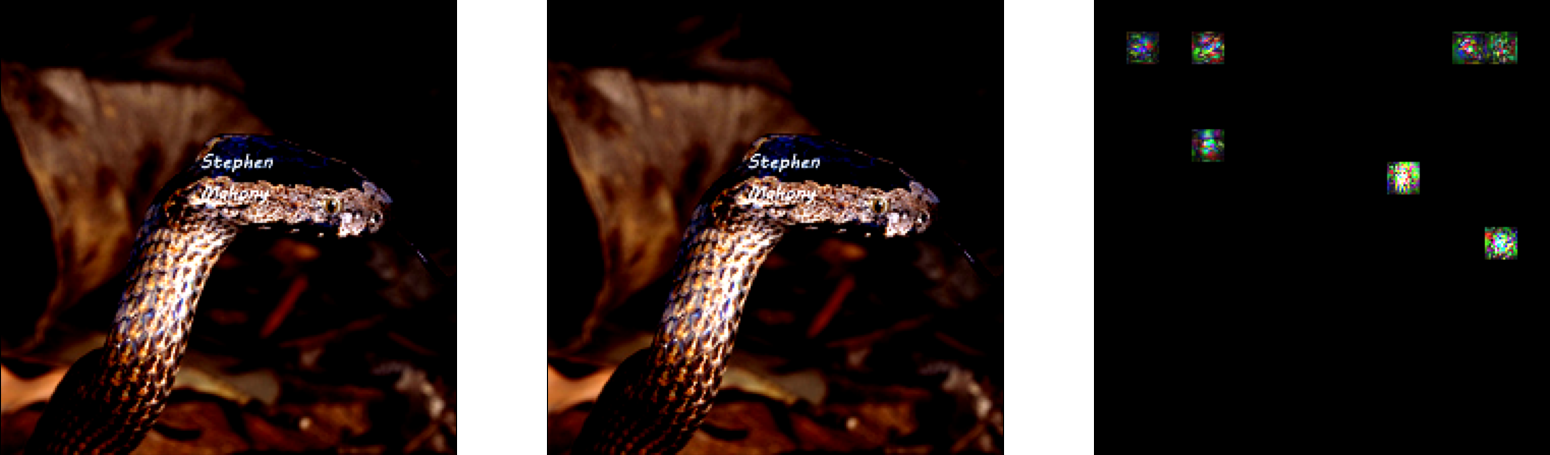} \\
  \raisebox{3em}{DeIT 224 (Distilled)} & \includegraphics[width=0.55\linewidth]{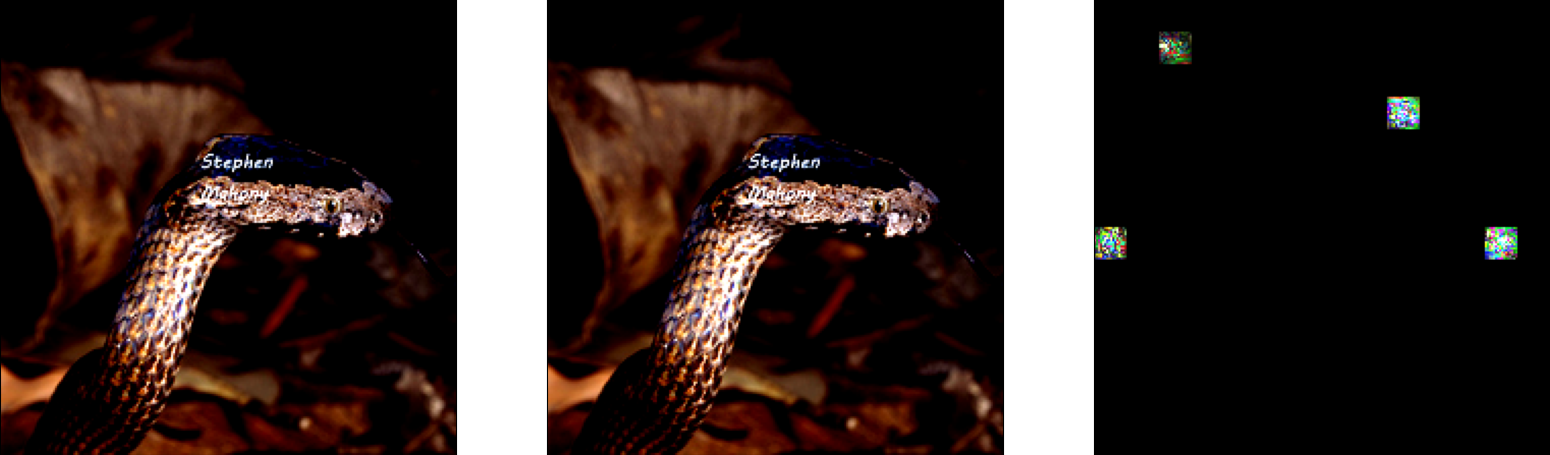}\\
\end{tabular}
\caption{\sl\textbf{Patch attacks on Transformers}: The attack images are generated with a fixed budget of $20$ patches. Note that the perturbations are imperceptible. The third column shows the perturbation brightened $10$ times.}
\label{fig:vis_blocksparse}
\end{figure}

\begin{table}
\begin{minipage}{0.5\linewidth}
\centering
\caption{\textbf{Robustness v/s Token Budget}}
\label{t:rob_tokenbudget}
\begin{tabular}{c c c c c c c}
    \toprule 
    \textbf{Model} & \multicolumn{6}{c}{Token Budget} \\
    \midrule 
    {} & 1 & 2 & 5 & 10 & 20 & 40  \\
    \midrule
    ViT-224 & 13.62 & 0.9 & 0.0 & 0.0 & 0.0 & 0.0  \\
    ViT-384 & 1.33 & 0.0 & 0.0 & 0.0 & 0.0 & 0.0  \\
    DeIT & 17.27 & 0.9 & 0.0 & 0.0 & 0.0 & 0.0  \\
    DeIT (Distilled) & 17.94 & 0.0 & 0.0 & 0.0 & 0.0 & 0.0  \\
    Resnet-101 & 49.50 & 32.22 &  8.64 & 1.66 & 0.33 & 0.0  \\
    Resnet-50 & 19.9 & 4.65 & 0.33 & 0.0 & 0.0 & 0.0 \\
    Wide-Resnet & 26.57 & 9.96 &  0.66 & 0.0 & 0.0  & 0.0 \\
    MLP-Mixer & 26.91 & 5.31 & 0.0 & 0.0 & 0.0 & 0.0 \\
    \bottomrule
\end{tabular}
\end{minipage}\hspace{20pt}
\begin{minipage}{0.5\linewidth}
\centering
\caption{\textbf{Robustness v/s varying patch sizes}}
\begin{tabular}{c c c c c}
\toprule
\textbf{Model} & \multicolumn{4}{c}{Attack patch sizes} \\
\midrule
{} & 1 & 4 & 8 & 16 \\
\midrule 
ViT-224 & 71.09  & 55.15 & 9.30 & 0.0  \\
ViT-384 & 68.77  & 31.89 & 0.06 & 0.0\\
DeIT & 78.40 & 68.77 & 8.31 & 0.0\\
DeIT-Distilled & 83.72 & 68.10 & 12.29 & 0.0 \\
Resnet-101d & 75.08 & 64.78 & 38.87 & 8.64 \\
Resnet-50 & 62.12 & 40.53 & 11.96 & 0.33 \\
Wide Resnet & 44.85 & 28.24 & 9.63 & 0.66 \\
MLP-Mixer & 76.41 & 54.49 & 17.61 & 5.31 \\
\bottomrule
\end{tabular}
\end{minipage}
\end{table}

\begin{figure*}
\begin{tabular}{c c}
\raisebox{5em}{ViT224} & \includegraphics[width=0.75\linewidth]{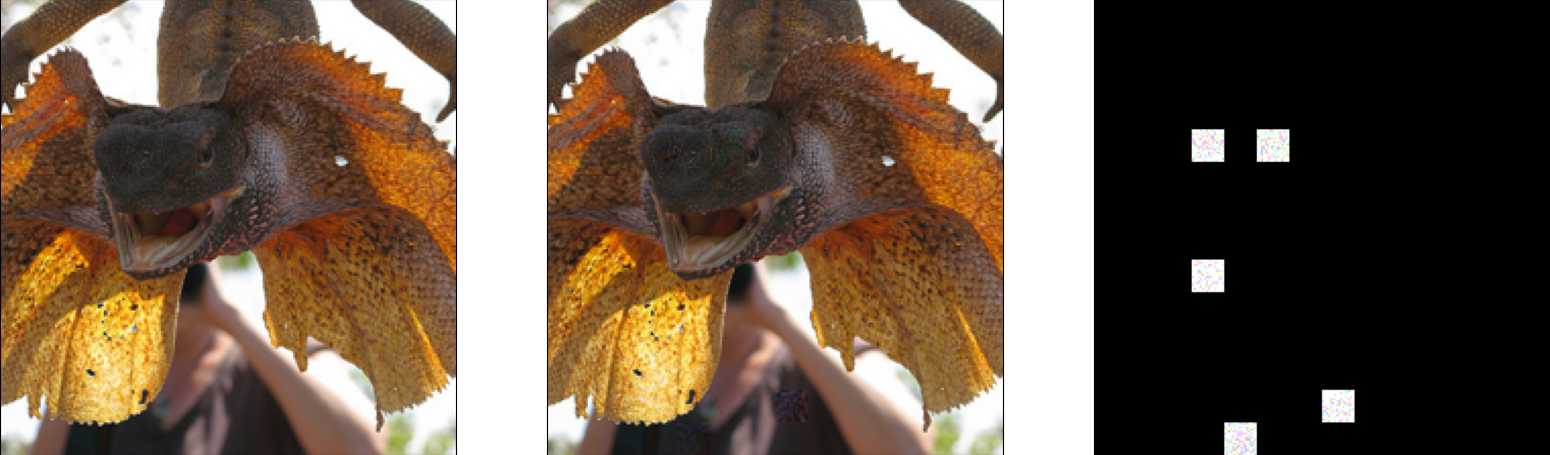} \\
\raisebox{5em}{DeIT} & \includegraphics[width=0.75\linewidth]{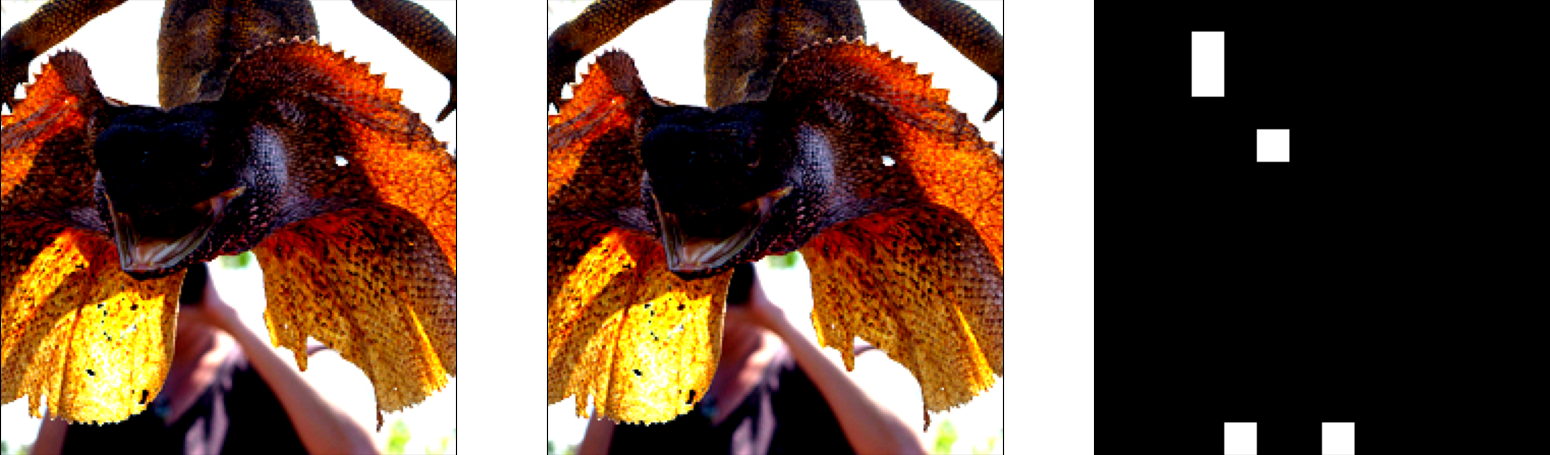} \\
\raisebox{5em}{DeiT-Distilled} & \includegraphics[width=0.75\linewidth]{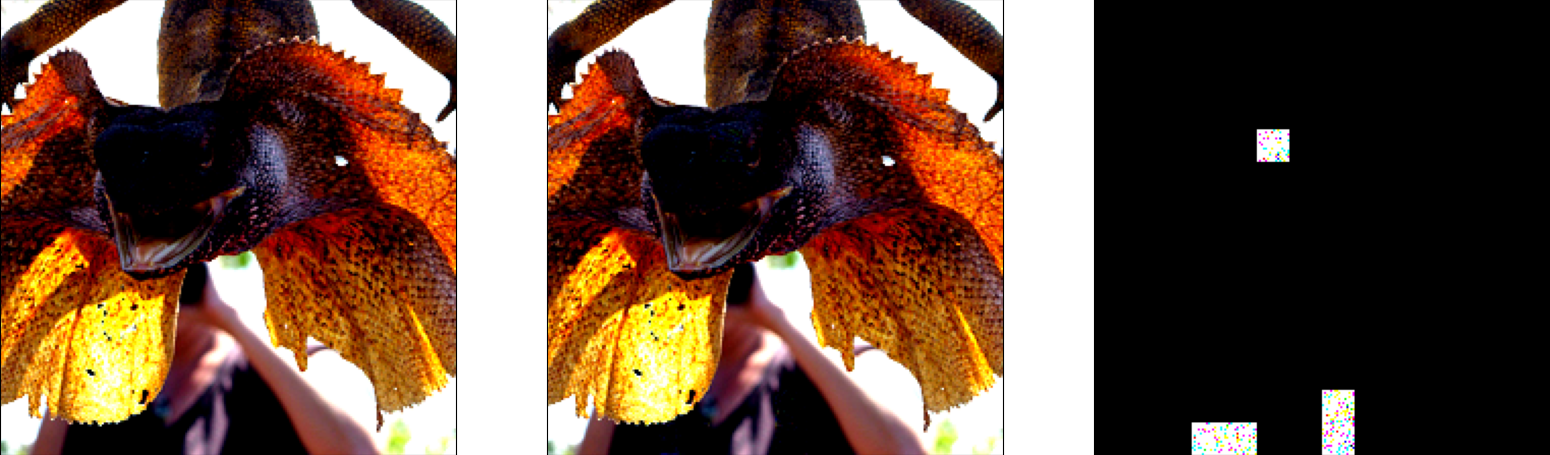} \\
\raisebox{5em}{Resnet-101} & \includegraphics[width=0.75\linewidth]{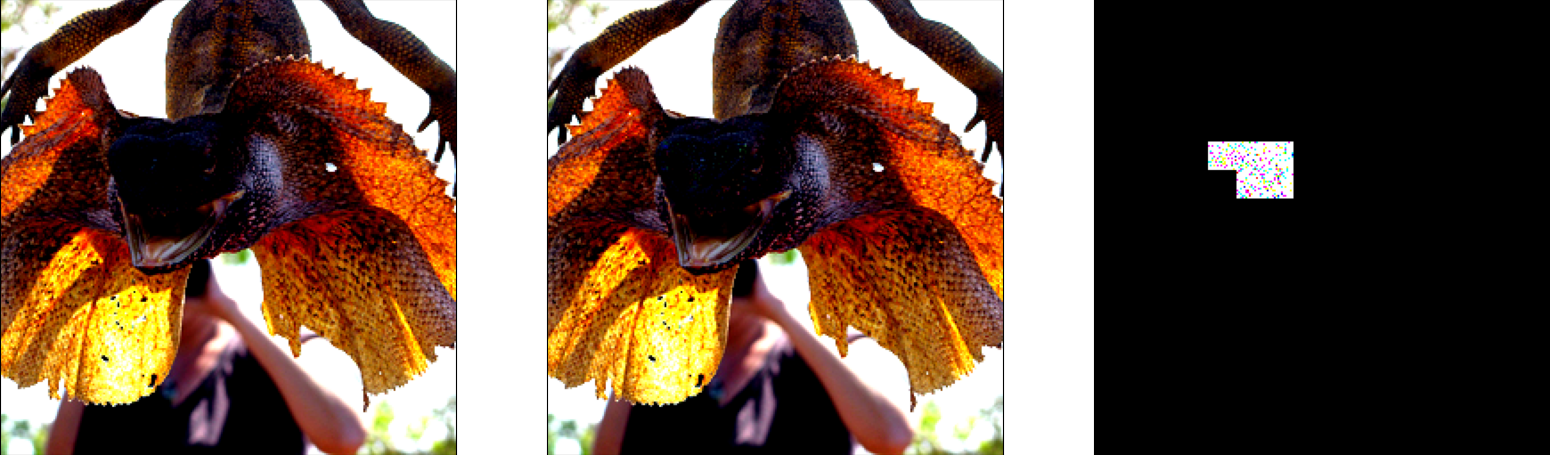} \\
\raisebox{5em}{Resnet-50} & \includegraphics[width=0.75\linewidth]{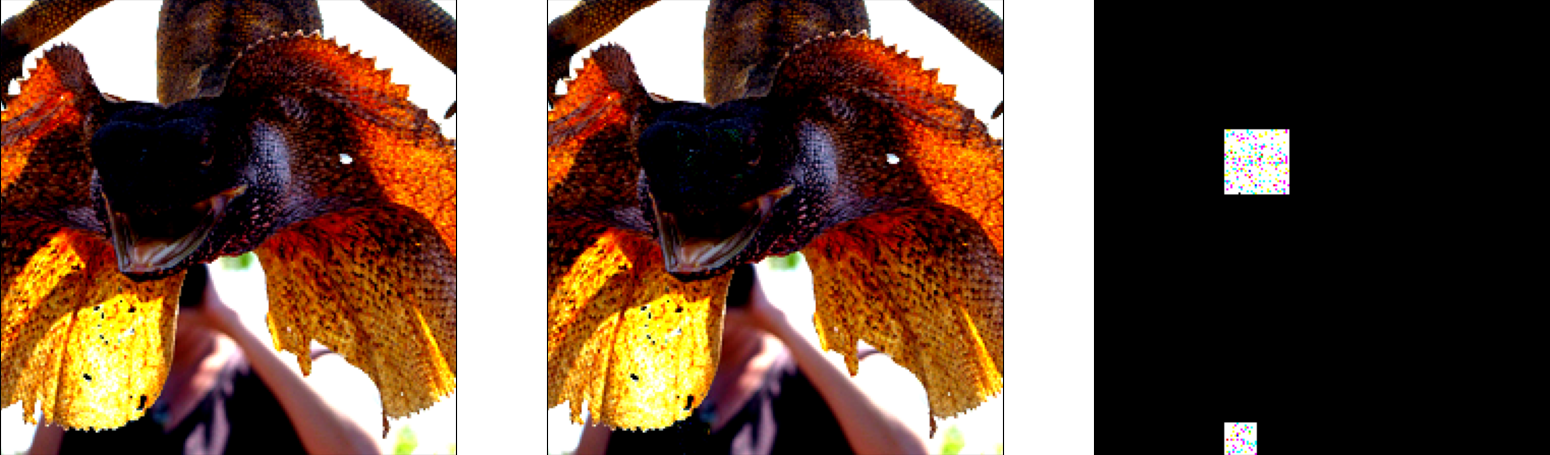} \\
\raisebox{5em}{MLP-Mixer} & \includegraphics[width=0.75\linewidth]{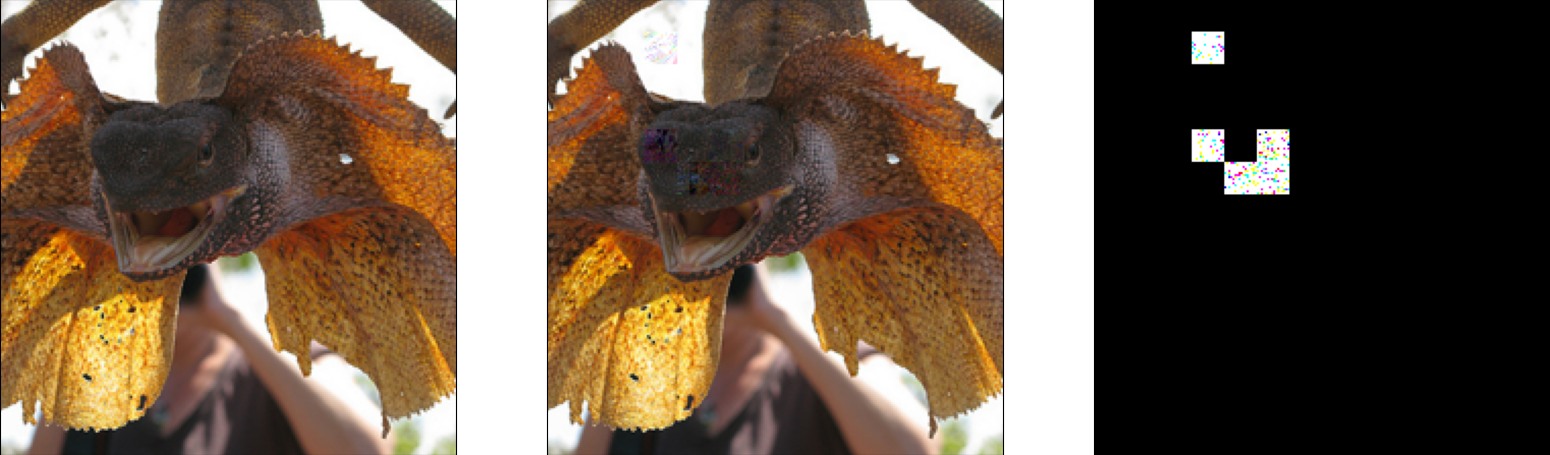} \\
\end{tabular}
\caption{\textbf{Examples of mixed norm attacks} Examples of successful $(5,(1/255)$ mixed norm attacks for various models}
\end{figure*}

\end{document}